\def\year{2022}\relax
\documentclass[letterpaper]{article} 
\usepackage{aaai22}  
\usepackage{times}  
\usepackage{helvet}  
\usepackage{courier}  
\usepackage[hyphens]{url}  
\usepackage{graphicx} 
\usepackage{natbib}  
\usepackage{caption} 
\DeclareCaptionStyle{ruled}{labelfont=normalfont,labelsep=colon,strut=off} 
\usepackage{algorithm}
\usepackage{algorithmic}
\usepackage{amsfonts}
\usepackage{amsmath}
\usepackage{booktabs}
\usepackage{multirow}
\usepackage{newfloat}
\usepackage{listings}
\floatstyle{ruled}
\newfloat{listing}{tb}{lst}{}
\floatname{listing}{Listing}
\title{Distribution-Free Federated Learning with Conformal Predictions}

\author {
    Charles Lu,\textsuperscript{\rm 1, 2, $\dagger$}
    Jayashree Kalpathy-Cramer\textsuperscript{\rm 1, \rm 2, \rm 3}
}
\affiliations {
    \textsuperscript{\rm 1} Athinoula A. Martinos Center for Biomedical Imaging, Massachusetts General Hospital, Charlestown, MA\\
    \textsuperscript{\rm 2} Center for Clinical Data Science, Massachusetts General Hospital and Brigham and Women's Hospital, Boston, MA\\
    \textsuperscript{\rm 3} Department of Radiology, Harvard Medical School, Boston, MA \\
    \textsuperscript{$\dagger$} clu@mgh.harvard.edu
}

\usepackage{bibentry}

\begin{document}

\maketitle

\begin{abstract}
    Federated learning has attracted considerable interest for collaborative machine learning in healthcare to leverage separate institutional datasets while maintaining patient privacy.
    However, additional challenges such as poor calibration and lack of interpretability may also hamper widespread deployment of federated models into clinical practice, leading to user distrust or misuse of ML tools in high-stakes clinical decision-making.
    In this paper, we propose to address these challenges by incorporating an adaptive conformal framework into federated learning to ensure distribution-free prediction sets that provide coverage guarantees. Importantly, these uncertainty estimates can be obtained without requiring any additional modifications to the model.
    Empirical results on the MedMNIST medical imaging benchmark demonstrate our federated method provides tighter coverage over local conformal predictions on 6 different medical imaging datasets for 2D and 3D multi-class classification tasks. 
    Furthermore, we correlate class entropy with prediction set size to assess task uncertainty.
\end{abstract}
%

\section{Introduction}
\label{sec:intro}
    Current machine learning techniques typically increase in predictive task performance when trained with more data.
    Because of their good predictive ability, convolutional neural networks (CNN) are often favored for many tasks in computer vision and medical imaging analysis.
    However, medical imaging datasets often have large dimensionality with a relatively limited number of labeled examples that make training CNNs more difficult~\cite{singh20203d,mri2019}.
    For low prevalence diseases, too few annotated cases may be available from a single healthcare institution to train a high performance model. 
    However, directly sharing private medical data with external institutions poses severe privacy and security risks.
    For this reason, federated learning (FL) has become an attractive paradigm to collaboratively develop CNN models while maintaining privacy of each institution's respective data~\cite{DBLP:journals/corr/abs-1911-06270}.
    
    Another considerable challenge of most standard CNN models is their lack of uncertainty estimation and poor calibration~\cite{guo2017calibration}.
    While powerful in predictive capability, deep CNNs are not generally considered interpretable or well calibrated, which can lead to higher levels of user distrust or frustration during high-stakes decision-making~\cite{bhatt2021uncertainty}.
    Having some way to guarantee statistical confidence of a prediction will be crucial in the adoption of ML tools into medical decision-making applications~\cite{ghassemi2019review}.
    
    Conformal predictions have been proposed as a general framework of providing distribution-free uncertainty in the form of prediction sets for classification and prediction intervals for regression \cite{10.5555/1062391,vovk2012conditional}.
    Additionally, prediction sets better match the intuition of clinical decision making by providing a set of possible conditions that may be used to rule in or rule out certain diseases similar to a differential diagnosis \cite{lu2021fair}.
    
    Our proposed method combines a federated training scheme with a adaptive conformal calibration during inference to combine the advantages of valid coverage with the increased performance and data privacy assurances of FL.
    To this end, we investigate the following three research questions (RQ):
    \begin{enumerate}
        \item How do conformal methods (adaptive prediction sets) compare to non-adaptive prediction sets for medical imaging classification tasks?
        \item Do a federated quantile estimates provide tighter prediction sets than quantiles calibrated only on the local healthcare institution?
        \item What is the relationship between conformal uncertainty (as measured by set size) and epistemic (model) uncertainty (as measured by class entropy)?
    \end{enumerate}

\section{Related Work}
\label{sec:related}
    Conformal inference was introduced by~\citet{10.5555/1062391} as a general framework to incorporate a rigorous notion of uncertainty into arbitrary machine learning models and have been adapted for prediction tasks including quantile regression, multi-label classification, and image segmentation~\cite{Romano2019ConformalizedQR, cauchois2020knowing, bates-rcps}.
    Conformal prediction can be seen as a frequentist approach to confidence estimation that provides marginal coverage bounds without needing to specify a prior distribution.
    
    Other approaches to confidence in CNN include Bayesian approximation and post-hoc probability calibration. 
    Bayesian neural networks permit confidence estimates using posterior probabilities and model uncertainty but usually impose strict assumptions on the modeling distribution, such as assuming a Gaussian distributed likelihood in Monte Carlo dropout~\cite{mc-dropout}.
    Other uncertainty methods are either model specific (such as Mondrian forests~\cite{pmlr-v51-lakshminarayanan16}) or require modification of the loss function during training~\cite{Corbire2019AddressingFP}. 
    
    The Softmax probability of CNN have been empirically shown lack proper calibration. Attempts to remedy improve these calibration issues in CNN have resulted in post-processing techniques such as temperature scaling and isotonic calibration~\cite{guo2017calibration}.
    
    FL has received considerable attention for privacy sensitive applications such as healthcare~\cite{digi-fed-learn}. 
    Previous works attempt to incorporate Bayesian uncertainty or address non-IID distributions with post-calibration in FL context~\cite{pmlr-v97-yurochkin19a, DBLP:journals/corr/abs-2106-05001}.
    
    Another work proposes distributed conformal predictions but does not consider a FL context with CNN for medical imaging applications~\cite{spjuth2018aggregating}.

\section{Methods}
\label{sec:methods}
    The conformal framework can be used to produce confidence prediction sets with any machine learning model that outputs a scoring function (e.g. deep neural networks, decision trees, and quantile regression).
    For CNN classifiers, this amounts to predicting a set of predictions using the Softmax probability (for multi-class classification) as the conformality score function. 
    One naive way of forming prediction sets is to simply sort the class scores for a given example and include all classes in the prediction set up to some confidence threshold. 
    For example, for a three-class prediction task with the Softmax class scores $[0.3, 0.1, 0.2, 0.4]$, our naive method would be output $\{3, 0, 2\}$ for a confidence level of $90\%$ and $\{3, 0\}$ for a confidence level of $70\%$.
    Unfortunately, this naive approach does not provide the property of \textit{coverage} --.
    simply, that the prediction set for any given example will contain the true class on average at some desired confidence level.
    To achieve a formal guarantee of coverage, we can instead reuse the validation data to calibrate our model to achieve valid conformal predictions during model inference.
    
    Formally, let $C(X)$ be a conformal prediction set, which is a subset of the label powerset, $\mathcal{P}$, and let $\alpha \in (0, 1)$ be a pre-chosen confidence level that determines the degree of guaranteed coverage on the joint distribution, $\{X_i, Y_i\}_{i=1}^{N + M}$, where $N$ is the number of examples in the calibration set and $M$ is the number of examples in the test set.
    Then, for some classifier that makes $\vert Y \vert$ class predictions using a monotonic scoring function, $S: \mathcal{X} \rightarrow \mathbb{R}^{\vert Y \vert}$, a prediction set can be created:  
    \begin{equation}
        C(X) = \{y \in Y \mid S(X)_y > 1 - \hat{q}\},
    \end{equation}
    where $S(X)_y$ is the score of the $y$th class and $\hat{q}$ is the score quantile (estimated using the calibration set $\{X_i, Y_i\}_{i=1}^N$) at the $1 - \alpha$ coverage level,
    \begin{equation}
        \hat{q} = \frac{\lceil(N + 1)(1 - \alpha)\rceil}{N},
    \end{equation}
    required to guarantee marginal coverage with a small finite sample correction, ~\cite{DBLP:journals/corr/abs-2009-14193}.
    If the data is drawn from an identical and exchangeable distribution (a weaker form of IID), then marginal coverage can be defined as satisfying the following property:
    \begin{equation}
        1 - \alpha \; \leq  \; P(Y \; \in \; C(X))
    \end{equation}
    This bound can be made near-tight through an appeal to symmetry (again assuming exchangeable random variables), which appears in several previous works~\cite{10.1007/3-540-36755-1_29, 10.5555/1062391, lei2017distributionfree, romano2020classification}.
    \begin{equation}
        1 - \alpha \; \leq  \; P(Y \; \in \; C(X)) \; \leq \; 1 - \alpha  + \frac{1}{M + N + 1}
    \end{equation}
    
    \begin{algorithm}[tb]
    \small
    \caption{Federated conformal prediction sets}
    \label{alg:fcps}
    \textbf{Input}: \\
        Federated model $S: X \times Y \rightarrow \mathbb{R}^{Y}$, \\
        $K$ validation sets $\{(X, Y)\}_{n=1}^N$, \\ 
        Test example $X_m$, \\ 
        Confidence level $1 - \alpha \in (0, 1)$, \\ 
        Quantile function $Q: \{\mathbb{R}\}_{i=1}^N \times [0, 1] \rightarrow \mathbb{R}$\\
        Sorting function $\pi: \mathcal{S} \rightarrow \mathcal{S}$\\
    \textbf{Output}: A confidence prediction set $C(X_m)$ \\
    \begin{algorithmic}[1]
        \STATE $\hat{q} \leftarrow 0$
        \FOR{$k \in \{1, 2, \ldots, K\}$}
            \FOR{$i \in \{1, 2, \ldots, N\}$}
                \STATE $s_{i} \leftarrow \sum_{j=1}^{y_j=y_i} \pi(S(x_i, y_i)$
            \ENDFOR
            \STATE $\hat{q} \leftarrow \hat{q} + Q(\{s_{i}\}^N_{i=1}, \lceil (1 - \alpha) (N + 1) \rceil)$ 
        \ENDFOR \
        \STATE \textbf{return }: $\{y \in Y : S(X_m)_y > \hat{q} / K \}$
    \end{algorithmic}
    \end{algorithm}
    
    Because this marginal coverage property does not impose any assumptions on the distribution (besides exchangeability) or score function (besides monotonicity), conformal prediction methods can be easily adapted for FL settings.
    Federated averaging (FedAvg) is a simple FL procedure that averages parameter weights of all local model (client side) to update the global model (server side) -- which is then distributed back to each local model -- after each training epoch~\cite{mcmahan2017communicationefficient}.
    As described in Algorithm \ref{alg:fcps}, we adapt split conformal predictions~\cite{romano2020classification} for FedAvg 
    While our conformal framework does not depend on a specific FL technique, we implement FedAvg for its simplicity in order to demonstrate a proof of concept.
    
    For more information on conformal predictions and distribution-free uncertainty, we refer the reader to the excellent tutorial by \citet{angelopoulos-gentle}.

\section{Experiments}
\label{sec:experiments}
    We compare three methods of forming prediction sets: a naive baseline, conformal predictions using local quantiles, and conformal predictions using federated quantiles.
    The naive baseline simply form predictions sets by adding sorting classes by predicted scores and adding elements to the set until the $1 - \alpha$ threshold is met.
    Because the naive baseline does not estimate the empirical quantile on a held-out calibrate set, we cannot expect it to satisfy the coverage guarantee described above. 
    
    For our experiments, we train a ResNet-18~\cite{He2016DeepRL} for 5 different random weight initializations to optimize a cross entropy loss on six datasets in the MedMNIST medical imaging dataset benchmark~\cite{medmnistv1} (BloodMNIST, DermaMNIST, PathMNIST, TissueMNIST, RetinaMNIST, OrganMNIST3D).
    The number of classes and overall class distribution for each dataset is shown in Figure \ref{fig:distribution}.  
    
    For each dataset, we split the original training, validation, and testing sets into four partitions to simulate different healthcare institutions to train our FL models.
    As conformal predictions are a ``post-proccessing'' step that do not depend on model training, the same trained models where used for all three methods of forming prediction sets.
    For local conformal prediction sets, we estimate the quantile using only the first institution's validation set as the calibration set.
    For federated conformal prediction sets, we calibrate the quantile using a simple average of quantiles from each individual institution's respective validation set.
    To further assess the performance of federated conformal and to tease out possible confounding of performance with larger calibration dataset sizes, we randomly shuffle 30\% of labels for three out of the four institutions' validation set while keeping the first institution's validation set ``clean''.
    All three methods were evaluated only on the first institution's test set.
    
\begin{figure}[htb]
        \centering
        \includegraphics[width=\linewidth]{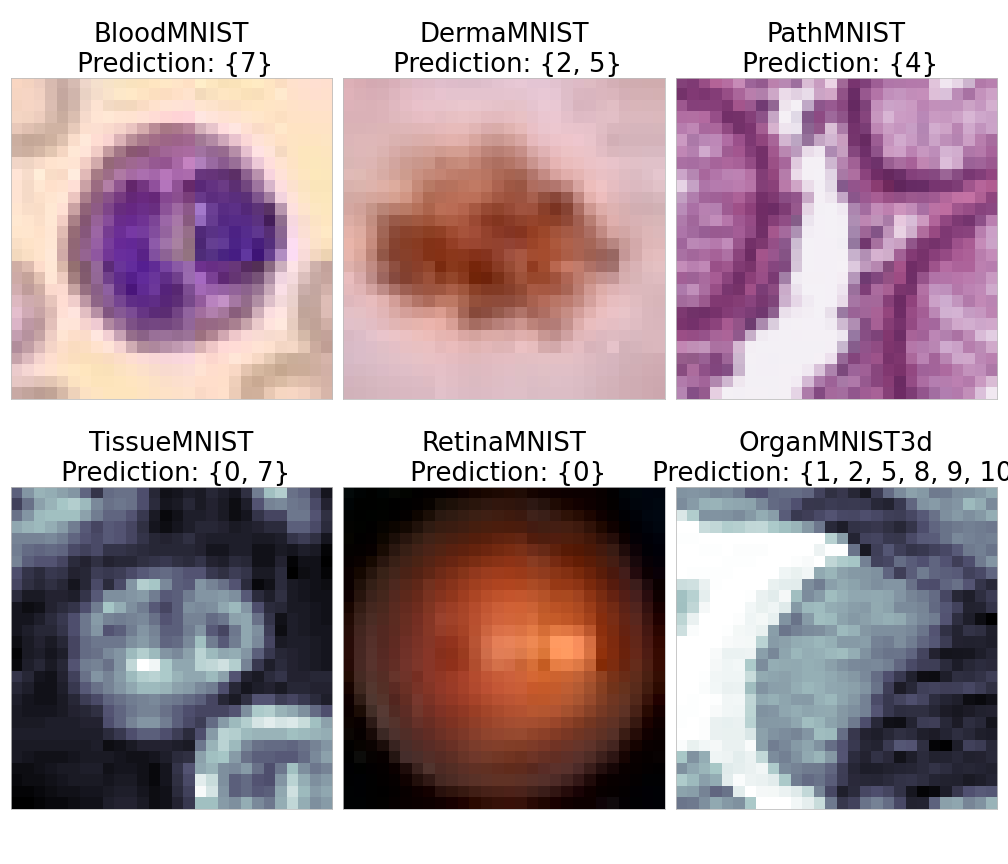}
        \caption{Image and corresponding prediction set at 90\% confidence level ( larger sets indicate higher uncertainty in the prediction task while smaller sets indicate more confidence in the prediction).}
        \label{fig:predictions}
    \end{figure}

\begin{figure}[htb]
        \centering
        \includegraphics[width=\linewidth]{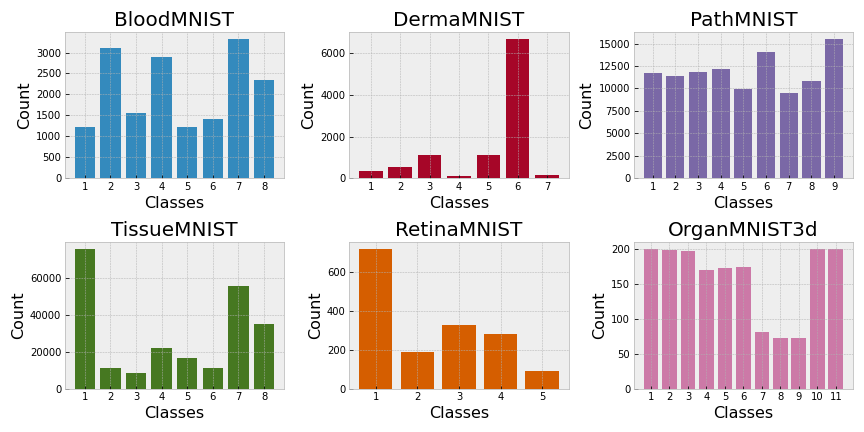}
        \caption{Varying class distribution of six MedMNIST benchmark datasets (see Figure \ref{fig:predictions} for an image from each dataset).}
        \label{fig:distribution}
    \end{figure}

    Images with the corresponding conformal prediction set is shown in Figure \ref{fig:predictions}.
    
    \begin{figure}[htb]
        \centering
        \includegraphics[width=\linewidth]{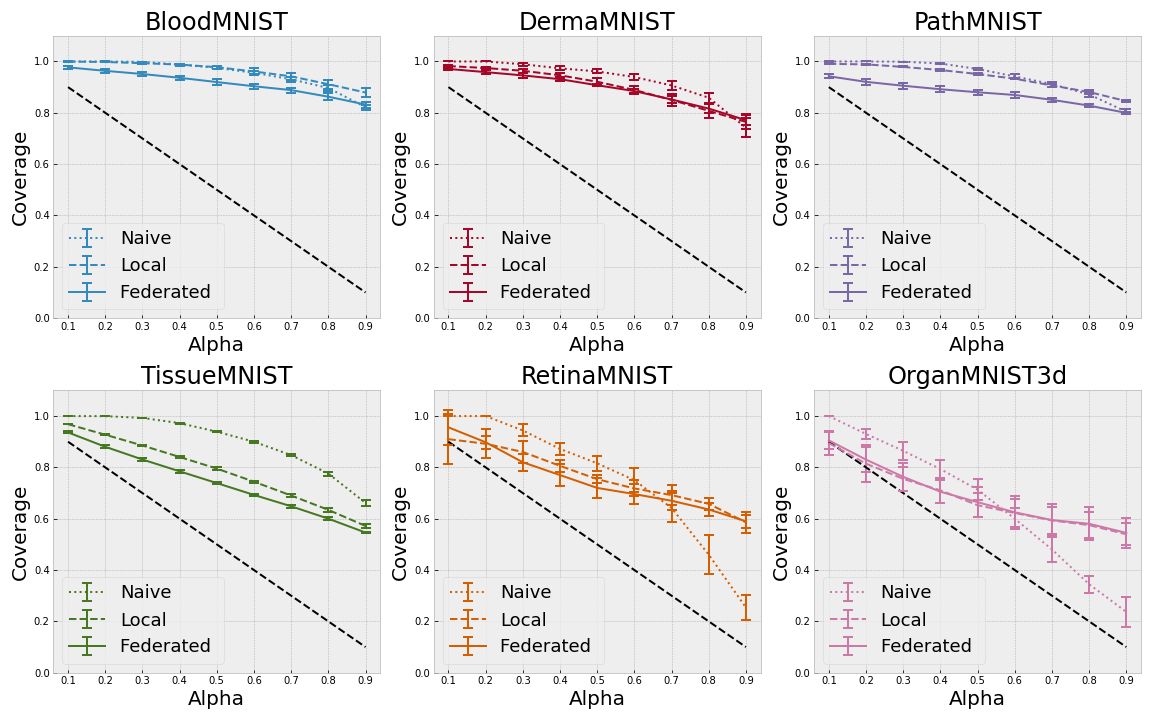}
        \caption{Coverage of naive, local conformal, and federated conformal methods by different levels of confidence ($1-\alpha$). Dashed line shows idealized coverage.}
        \label{fig:coverage}
    \end{figure}

    \begin{figure}[htb]
        \centering
        \includegraphics[width=\linewidth]{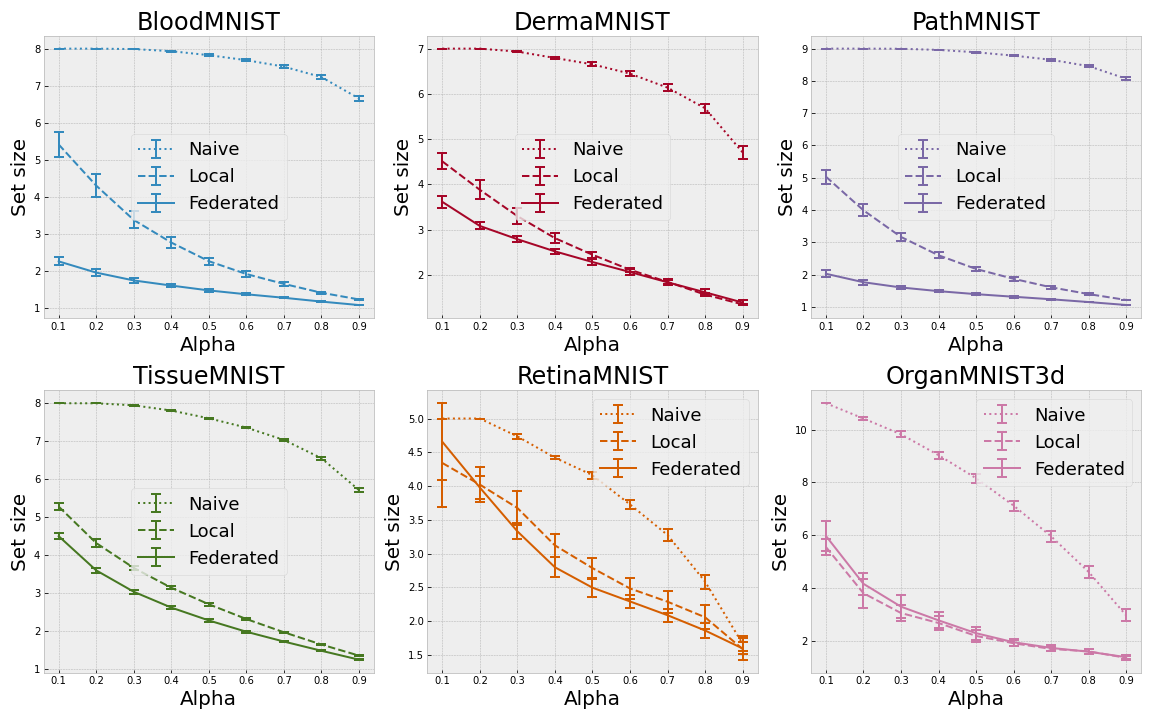}
        \caption{Average size of prediction sets of naive, local conformal, and federated conformal methods by different levels of confidence ($1-\alpha$).}
        \label{fig:cardinality}
    \end{figure}
    
    
    Additionally, we wish to study the relationship between the difficulty of predicting a specific case (independent to the number of classes) with its corresponding uncertainty estimate (as determined by the number of elements in the prediction set).
    To this end, we hold out 10\% of the training set from each dataset to train an auxiliary classifier, which can then be used to correlate the maximum Softmax score with the class-wise entropy, $H(X_i) = -\sum_{y \in Y} S(X_i)_y \cdot \log S(X_i)_y$, for the rest of the dataset.
    Therefore, we expect entropy to be indicative of model uncertainty and positively correlated with the size of a conformal prediction set.
    
    \begin{table}[htb]
    \tiny
    \centering
    \begin{tabular}{|c|c|c|c|c|c|}
    \hline
    \textbf{Dataset} & \textbf{Classes} & \textbf{Examples} &  \textbf{Method} & \textbf{Coverage} & \textbf{Cardinality}\\
    \hline
    \multirow{3}*{Blood} & \multirow{3}*{9} & \multirow{3}*{17,092}
    & Naive & 100\% $\pm$ 0\% & 8 $\pm$ 0 \\
    & & & Local & 100\% $\pm$ 0\% & 5.4 $\pm$ 0.3 \\
    & & & Federated & 97.7\% $\pm$ 0.7\% & 2.3 $\pm$ 0.1 \\
    \hline
    \multirow{3}*{Derma} & \multirow{3}*{7} & \multirow{3}*{10,015}
    & Naive &  100\% $\pm$ 0\% & 7 $\pm$ 0 \\
    & & & Local &  98.2\% $\pm$ 0.5\% & 4.5 $\pm$ 0.2 \\
    & & & Federated &  97.1\% $\pm$ 0.4\% & 3.6 $\pm$ 0.1 \\
    \hline
    \multirow{3}*{Path} & \multirow{3}*{9} & \multirow{3}*{107,180}
    & Naive &  100\% $\pm$ 0\% & 8 $\pm$ 0 \\
    & & & Local &  99.1\% $\pm$ 0.2\% & 5 $\pm$ 0.2 \\
    & & & Federated &  94.2\% $\pm$ 0.8\% & 2 $\pm$ 0.1 \\
    \hline
    \multirow{3}*{Tissue} & \multirow{3}*{8} & \multirow{3}*{236,386}
    & Naive &  100\% $\pm$ 0\% & 8 $\pm$ 0 \\
    & & & Local &  96.8\% $\pm$ 0.1\% & 5.3 $\pm$ 0.1 \\
    & & & Federated &  93.6\% $\pm$ 0.4\% & 4.5 $\pm$ 0.1 \\
    \hline
    \multirow{3}*{Retina} & \multirow{3}*{5} & \multirow{3}*{1,600}
    & Naive &  100\% $\pm$ 0\% & 5 $\pm$ 0 \\
    & & & Local &  91\% $\pm$ 9.8\% & 4.3 $\pm$ 0.7 \\
    & & & Federated &  95.6\% $\pm$ 6.9\% & 4.7 $\pm$ 0.6 \\
    \hline
    \multirow{3}*{Organ3d} & \multirow{3}*{11} & \multirow{3}*{1,743}
    & Naive &  100\% $\pm$ 0\% & 11 $\pm$ 0 \\
    & & & Local &  89.5\% $\pm$ 4.5\% & 5.6 $\pm$ 0.3 \\
    & & & Federated &  90.5\% $\pm$ 3.5\% & 6 $\pm$ 0.6 \\
    \hline
    \end{tabular}
    \caption{Summary of experimental results for each dataset at setting of 90\% coverage ($\alpha=0.1$).}
    \label{table:summary}
    \end{table}
    
    \begin{figure}[htb]
        \centering
        \includegraphics[width=\linewidth]{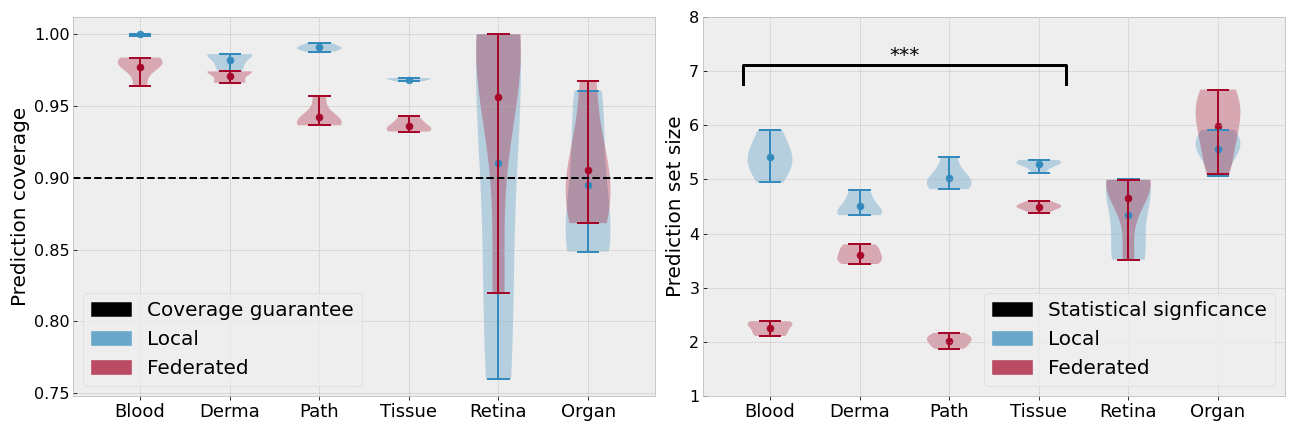}
        \caption{(Left) figure shows average test coverage at $\alpha=0.9$ of local conformal and federated conformal methods across six MedMNIST classification tasks; (right) figure shows federated prediction sets having lower average cardinality than local prediction sets at $p<0.001$ significance for BloodMNIST, DermaMNIST, PathMNIST, and TissueMNIST datasets}
        \label{fig:violin}
    \end{figure}
    
    \begin{figure}[htb]
        \centering
        \includegraphics[width=\linewidth]{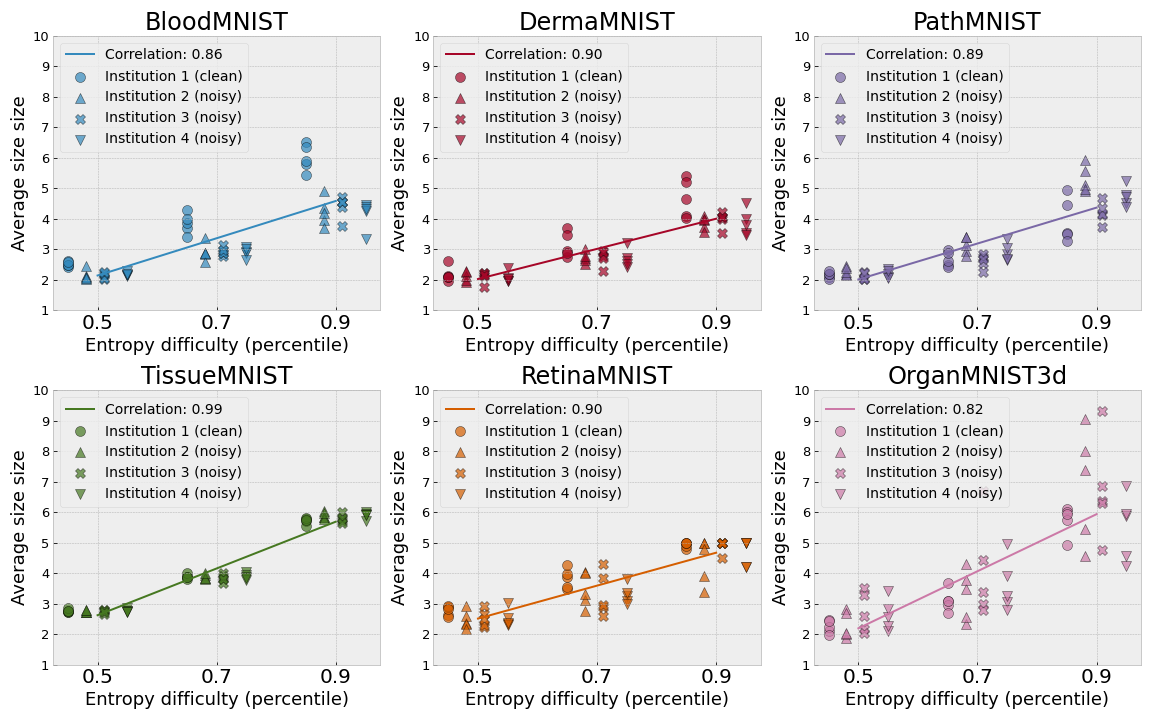}
        \caption{Correlation of local conformal set size and entropy percentile for each of the four institutions (1 clean; 3 trained with 30\% random labels).}
        \label{fig:correlation}
    \end{figure}
    
\section{Discussion}
\label{sec:discussion}
    We can evaluate and compare different methods of predictions sets by their cardinality (average number of elements in a prediction set) and coverage (probability that the true class is an element of the prediction set).
    Ideally, we would also like prediction sets to be ``adaptive'' to communicate model uncertainty at an instance level (difficult cases should have larger confidence sets while easier cases should have fewer elements in their prediction sets) while still satisfying marginal coverage guarantees. 
    From the results of our experiments, which are summarized in Table \ref{table:summary}, we can immediately see the naive baseline prediction sets as exceptionally conservative in coverage across all six datasets.
    In order words, naively forming predictions sets using raw Softmax ``probabilities'' simply outputs all possible classes for each prediction for 100\% coverage but useless for any kind of prediction task.
    Our results validate previous studies on overconfident Softmax scores in deep neural networks~\cite{guo2017calibration, hendrycks2018baseline}.
    While post-processing calibration techniques may produce more useful confidence sets, they do not offer the statistical guarantees provided by conformal prediction methods. 
    
    Next, we compare the two conformal prediction methods: prediction sets calibrated with a federated quantile and prediction sets calibrated with a local quantile.
    As shown in Table \ref{table:summary} and \ref{fig:coverage}, while all methods satisfy coverage properties (for example at a $\alpha$ threshold of 90\%), both conformal methods have much tighter confidence sets in terms of empirical cardinality compared to the naive baseline (see Figure \ref{fig:cardinality}).
    Additionally, federated conformal predictions have lower cardinality than local conformal predictions across MedMNIST datasets for at most $\alpha$ levels (see Figure \ref{fig:cardinality}) -- even when label noise was introduced to 30\% of the validation set used to estimate the federated conformal prediction quantile.
    Notably, our federated method performs remarkably well on both the BloodMNIST and PathMNIST datasets, which both have mostly equal distribution of classes (see Figure \ref{fig:distribution}), achieving average prediction set sizes of $2.3 \pm 0.1$ and $2 \pm 0.1$, respectively, and having a much lower rate of set size with at lower $\alpha$ thresholds compared to local conformal prediction sets (see \ref{fig:cardinality}).
    However, on both RetinaMNIST and OrganMNIST3d datasets, prediction sets formed using federated quantile estimate show a similar average set size as prediction sets formed with a local quantile estimate; however, both dataset also have (between 10-200 times) fewer cases than the other datasets.
    As shown in Figure \ref{fig:violin}, both these smaller datasets have much larger variance bands for both coverage and prediction set size.
    Therefore, we conclude that conformal methods outperform the naive baseline of forming prediction sets (RQ1) and that federated conformal prediction sets have smaller set sizes than local conformal prediction sets (RQ2) for datasets with a sufficient number of examples.
    
    To answer RQ3, we plot each of the four institutions at different three different class entropy percentiles (50\%, 75\%, and 90\%) each dataset (see Figure \ref{fig:correlation}).
    We confirm a positive relationship between between cardinality and epistemic model uncertainty, and also observe that conformal uncertainty to be robust to moderate levels of label corruption in the calibration set used to estimate the empirical quantile.
    
\section{Conclusion}
\label{sec:conclusion}
    We proposed a novel approach to combine FL with conformal prediction sets and empirically evaluated our method on several benchmark medical imaging datasets (MedMNIST) to demonstrate feasibility of conformal uncertainty in healthcare. 
    Additionally, we find a intuitive correspondence between conformal uncertainty and prediction task difficulty as measured by predictive class entropy.
    Combining the benefits of federated training with conformal inference, our distribution-free method of uncertainty quantification is a simple and flexible technique that requires no direct modification to model training while retaining wide applicability for collaboratively developing FL models among healthcare institutions.

\bibliography{aaai22.bib}

\end{document}